\documentclass{article}

\PassOptionsToPackage{numbers, compress}{natbib}

\usepackage[final]{neurips_2023}

\usepackage[utf8]{inputenc} 
\usepackage[T1]{fontenc}    
\usepackage{hyperref}       
\usepackage{url}            
\usepackage{booktabs}       
\usepackage{amsfonts}       
\usepackage{nicefrac}       
\usepackage{microtype}      
\usepackage{xcolor}         

\usepackage[pdftex]{graphicx}
\usepackage{amsmath}
\usepackage{appendix}
\DeclareMathOperator*{\argmin}{argmin}

\title{Minecraft-ify: Minecraft Style Image Generation with Text-guided Image Editing for In-Game Application}

\author{
  Bumsoo Kim$^{1, 2}$, Sanghyun Byun$^3$, Yonghoon Jung$^3$, \\ \textbf{Wonseop Shin$^3$, Sareer UI Amin$^4$, Sanghyun Seo\thanks{Corresponding author}}  \\
  $^{1,*}$School of Art and Technology, Chung-Ang University \\
  $^{2}$VIVE STUDIOS\\
  $^3$GSAIM, Chung-Ang University \\
  $^4$Department of Computer Science and Engineering, Chung-Ang University \\
  $^{1,2}$\texttt{bumsookim00@gmail.com}, \texttt{\{$^3$egoist12276, $^3$dydgns2017}, \\ \texttt{$^3$wonseop218, $^4$sarrer2021, $^*$sanghyun\}@cau.ac.kr} \\
}

\begin{document}

\maketitle

\begin{abstract}
In this paper, we first present the character texture generation system \textit{Minecraft-ify}, specified to Minecraft video game toward in-game application. Ours can generate face-focused image for texture mapping tailored to 3D virtual character having cube manifold. While existing projects or works only generate texture, proposed system can inverse the user-provided real image, or generate average/random appearance from learned distribution. Moreover, it can be manipulated with text-guidance using StyleGAN and StyleCLIP. These features provide a more extended user experience with enlarged freedom as a user-friendly AI-tool. Project page can be found at \textcolor{magenta}{\url{https://gh-bumsookim.github.io/Minecraft-ify/}}
\end{abstract}

\begin{figure}[!htbp]
\centering
    \includegraphics[width=\linewidth]{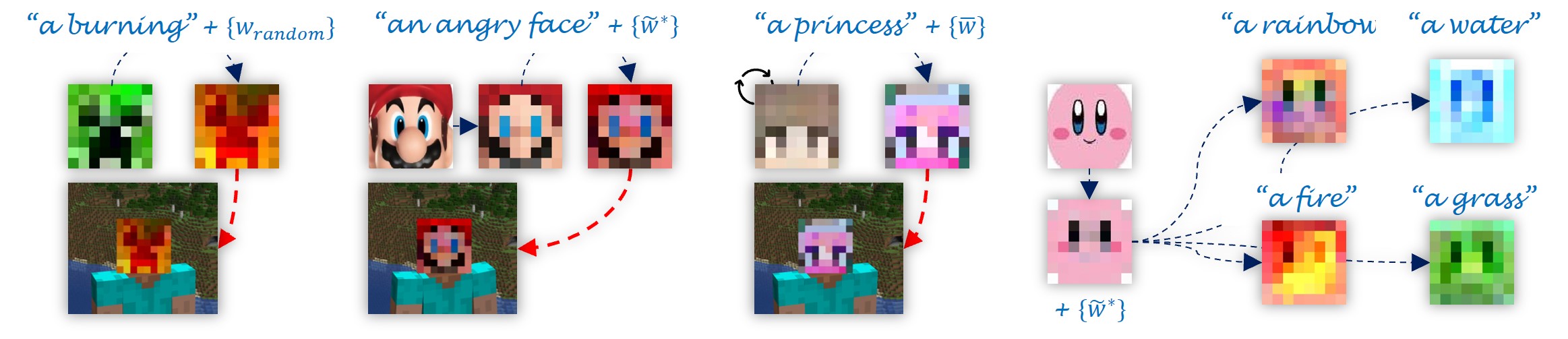}
    \hfil
\caption{Rendered 3D character in Minecraft-World using our generated frontal character texture.}	
\label{fig:results}
\end{figure}

\section{Introduction}

Non-photorealistic image generative models with editability leads AI-based creation tools for comic book, animation and video game. Recently, for specific domain with applicability, generative model have been tailored with data-centric approach \cite{Hao2021GANcraftU3, Wu2022MakeYO, Zhao2023ZeroShotTT, Pinkney2020ResolutionDG, Back2022WebtoonMeAD, Li2023ParsingConditionedAT}. In this paper, we present creation tool, specified the 3D character texture of Mincraft-World\footnote{https://www.minecraft.net/} based on StyleGAN \cite{Karras2018ASG, Karras2019AnalyzingAI} and StyleCLIP \cite{patashnik2021styleclip} including text-guided manipulation. With elaborately refined large Minecraft-World character texture dataset, game player can generate the frontal face texture of 3D character and manipulate it via text description with extended user experience and freedom.

\section{Method}
\label{Method}

\begin{figure}[!htbp]
\centering
    \includegraphics[width=\linewidth]{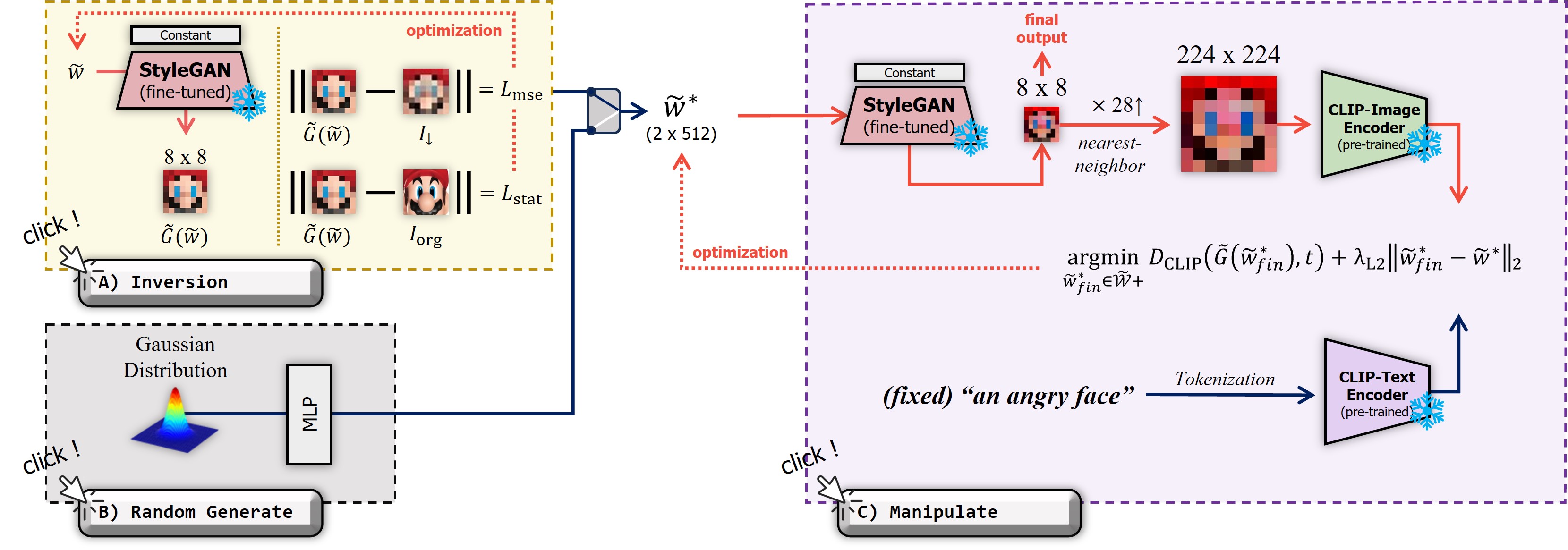}
    \hfil
\caption{Overview of our \textit{Minecraft-ify} system.}	
\label{fig:abstract}    
\end{figure}

Our proposed system aims to generate and manipulate Minecraft-World character image having texture format. From that, we can provide wider user freedom for character creation with two paths: A) inverse the user-provided real images, or B) generate the frontal texture from learned distribution. Finally, they can manipulate generated image with text description. For inversion, our inversion objective, was originally proposed from Image2StyleGAN \cite{Abdal2019Image2StyleGANHT}, designed with simple modification:

\begin{equation}
\label{eq:inversion}
\argmin_{\tilde{w} \in \mathcal{\tilde{W}+}} \frac {\lambda_{\text{mse}}} {N} \left \| \tilde{G}(\tilde{w})-I_{\downarrow} \right \|^2_2 + \lambda_{\text{stat}}L_{\text{stat}}(\tilde{G}(\tilde{w}), I_{\text{org}}),
\end{equation}

where $\tilde{G}(\cdot)$ is fined-tuned generator trained in preprocess with our large dataset which output the 8 by 8 image, $\tilde{w}\in\mathbb{R}^{2 \times 512}$ is limited latent vector in limited space $\mathcal{\tilde{W}+}$ for specified Minecraft-World texture, $I_{\downarrow}$ is downsampled real image with same size as $\tilde{G}(\tilde{w})$ and $L_{\text{stat}}$ is statistics loss obtained by:

\begin{equation}
\label{eq:loss_stat}
L_{\text{stat}}(\tilde{G}(\tilde{w}), I_{\text{org}}) = \frac{1}{3} \sum_{c \in \textbf{\{R,G,B\}}} \left (  | \mu_c(\tilde{G}(\tilde{w}) - \mu_c(I_{\text{org}}) | + | \sigma_c(\tilde{G}(\tilde{w}) - \sigma_c(I_{\text{org}}) | \right ),
\end{equation}

where $\mu_c$ and $\sigma_c$ are mean and standard deviation of $c$ channel, respectively. With $L_{\text{stat}}$, we explicitly force the generated texture to have similar image statistics with real image $I_{\text{org}}$ inspired by \cite{afifi2021histogan}. After inversion, we apply the StyleCLIP \cite{patashnik2021styleclip} via text using latent optimization method without identity loss:

\begin{equation}
\label{eq:clip}
\argmin_{\tilde{w}^*_{fin} \in \mathcal{\tilde{W}+}} D_{\text{CLIP}}(\tilde{G}(\tilde{w}^*_{fin}), t) + \lambda_{\text{L2}} \left \| \tilde{w}^*_{fin}-\tilde{w}^* \right \|_2,
\end{equation}

where $\tilde{w}^*$ is fixed vector obtained by inversion process,  $D_{\text{CLIP}}$ output the similarity between image and text using CLIP \cite{Radford2021LearningTV} image-, text-encoder and $t$ is tokenized vector from text description. From Eq. \ref{eq:clip}, we can finalize the manipulation process for in-game texture generation and editing through StyleCLIP-based \cite{patashnik2021styleclip} optimized vector $\tilde{w}^*_{fin}$ as $\tilde{G}(\tilde{w}^*_{fin})$. Player also can utilize average vector $\bar{w}$ or random vector $w_{\text{random}}$ instead of inversed vector $\tilde{w}^*$ in Eq. \ref{eq:clip} without considering real image input.

\section{Conclusion}

To generate and manipulate the Minecraft-World texture toward in-game application, we proposed \textit{Minecraft-ify} that can fully support the functions for enhanced user-freedom as user-friendly AI-tool using StyleGAN \cite{Karras2018ASG, Karras2019AnalyzingAI} and StyleCLIP \cite{patashnik2021styleclip}. From experimental results, we demonstrated that the text-guided manipulation can enough provide semantically plausible appearance although it was derived from user-wanted real sample by inversion. Additionally, we also showed that user can generate seamless random or average appearance texture from the learned distribution without considering the input images.

\section{Ethical Implications}

Our large dataset originally obtained from \texttt{\href{https://www.kaggle.com/datasets/sha2048/minecraft-skin-dataset?select=Skins}{here}} using Public Domain license. Our system generate the image via text with CLIP \cite{patashnik2021styleclip}. CLIP is known to have unwanted data-bias issues by training dataset. Thus, it is important that the user do not use this work for generating harmful or unpleasant things. Note that this work is proposed for entertainment purposes only to easily create diverse character texture to enrich the in-game play experience.

\section{Acknowledgement}

This research was supported by Culture, Sports and Tourism R\&D Program through the Korea Creative Content Agency(KOCCA) grant funded by the Ministry of Culture, Sports and Tourism(MCST) in 2023 (Project Name: Development of digital abusing detection and management technology for a safe Metaverse service, Project Number: RS-2023-00227686, Contribution Rate: 50\%) and the National Research Foundation of Korea (NRF) grant funded by the Korean government (MSIT) (No.2022R1A2C1004657, Contribution Rate: 50\%).

\bibliography{bibtext.bib}

\begin{thebibliography}{10}

\bibitem{Abdal2019Image2StyleGANHT}
R.~Abdal, Y.~Qin, and P.~Wonka.
\newblock Image2stylegan: How to embed images into the stylegan latent space?
\newblock {\em 2019 IEEE/CVF International Conference on Computer Vision (ICCV)}, pages 4431--4440, 2019.

\bibitem{afifi2021histogan}
M.~Afifi, M.~A. Brubaker, and M.~S. Brown.
\newblock Histogan: Controlling colors of gan-generated and real images via color histograms.
\newblock In {\em Proceedings of the IEEE/CVF conference on computer vision and pattern recognition}, pages 7941--7950, 2021.

\bibitem{Back2022WebtoonMeAD}
J.~Back, S.~Kim, and N.~Ahn.
\newblock Webtoonme: A data-centric approach for full-body portrait stylization.
\newblock {\em SIGGRAPH Asia 2022 Technical Communications}, 2022.

\bibitem{Hao2021GANcraftU3}
Z.~Hao, A.~Mallya, S.~J. Belongie, and M.-Y. Liu.
\newblock Gancraft: Unsupervised 3d neural rendering of minecraft worlds.
\newblock {\em 2021 IEEE/CVF International Conference on Computer Vision (ICCV)}, pages 14052--14062, 2021.

\bibitem{Karras2018ASG}
T.~Karras, S.~Laine, and T.~Aila.
\newblock A style-based generator architecture for generative adversarial networks.
\newblock {\em 2019 IEEE/CVF Conference on Computer Vision and Pattern Recognition (CVPR)}, pages 4396--4405, 2018.

\bibitem{Karras2019AnalyzingAI}
T.~Karras, S.~Laine, M.~Aittala, J.~Hellsten, J.~Lehtinen, and T.~Aila.
\newblock Analyzing and improving the image quality of stylegan.
\newblock {\em 2020 IEEE/CVF Conference on Computer Vision and Pattern Recognition (CVPR)}, pages 8107--8116, 2019.

\bibitem{Li2023ParsingConditionedAT}
Z.~Li, Y.~Xu, N.~Zhao, Y.~Zhou, Y.~Liu, D.~Lin, and S.~He.
\newblock Parsing-conditioned anime translation: A new dataset and method.
\newblock {\em ACM Transactions on Graphics}, 42:1 -- 14, 2023.

\bibitem{patashnik2021styleclip}
O.~Patashnik, Z.~Wu, E.~Shechtman, D.~Cohen-Or, and D.~Lischinski.
\newblock Styleclip: Text-driven manipulation of stylegan imagery.
\newblock In {\em Proceedings of the IEEE/CVF International Conference on Computer Vision}, pages 2085--2094, 2021.

\bibitem{Pinkney2020ResolutionDG}
J.~N.~M. Pinkney and D.~Adler.
\newblock Resolution dependent gan interpolation for controllable image synthesis between domains.
\newblock {\em ArXiv}, abs/2010.05334, 2020.

\bibitem{Radford2021LearningTV}
A.~Radford, J.~W. Kim, C.~Hallacy, A.~Ramesh, G.~Goh, S.~Agarwal, G.~Sastry, A.~Askell, P.~Mishkin, J.~Clark, G.~Krueger, and I.~Sutskever.
\newblock Learning transferable visual models from natural language supervision.
\newblock In {\em International Conference on Machine Learning}, 2021.

\bibitem{Wu2022MakeYO}
Z.~Wu, L.~Chai, N.~Zhao, B.~Deng, Y.~Liu, Q.~Wen, J.~Wang, and S.~He.
\newblock Make your own sprites.
\newblock {\em ACM Transactions on Graphics (TOG)}, 41:1 -- 16, 2022.

\bibitem{Zhao2023ZeroShotTT}
R.~Zhao, W.~Li, Z.~Hu, L.~Li, Z.~Zou, Z.~X. Shi, and C.~Fan.
\newblock Zero-shot text-to-parameter translation for game character auto-creation.
\newblock {\em 2023 IEEE/CVF Conference on Computer Vision and Pattern Recognition (CVPR)}, pages 21013--21023, 2023.

\end{thebibliography}
\bibliographystyle{abbrv} 

\appendix
\appendixpage

\section{StyleGAN fine-tuning} \label{appendix}

Before GAN inversion and CLIP-based optimization process, generator is fine-tuned with face texture dataset. Since our output has 8 by 8 image, the partial convolution layers are learnable in training. Thereby, latent vector also include first two coarse-level elements $\tilde{w}\in\mathbb{R}^{2 \times 512}$.

In training, fine tuning from FFHQ-weight converge more fast than weight initialization (\textit{i.e.}, from scratch), and we used 1024, 512 batch size for 4 by 4 and 8 by 8 output, respectively. For 20K iterations, it took about 6 hours under 1 NVIDIA RTX 3090 24GB.

Generator architecture is based on StyleGAN1\footnote{https://github.com/SiskonEmilia/StyleGAN-PyTorch} since we can not find any difference between StyleGAN1 and StyleGAN2 outputs. In our knowledge, this is because output pixel has low representation capability with partially used convolution layers against overall StyleGAN.

\section{Dataset refinement} \label{appendix}

\begin{figure}[!htbp]
\centering
    \includegraphics[width=\linewidth]{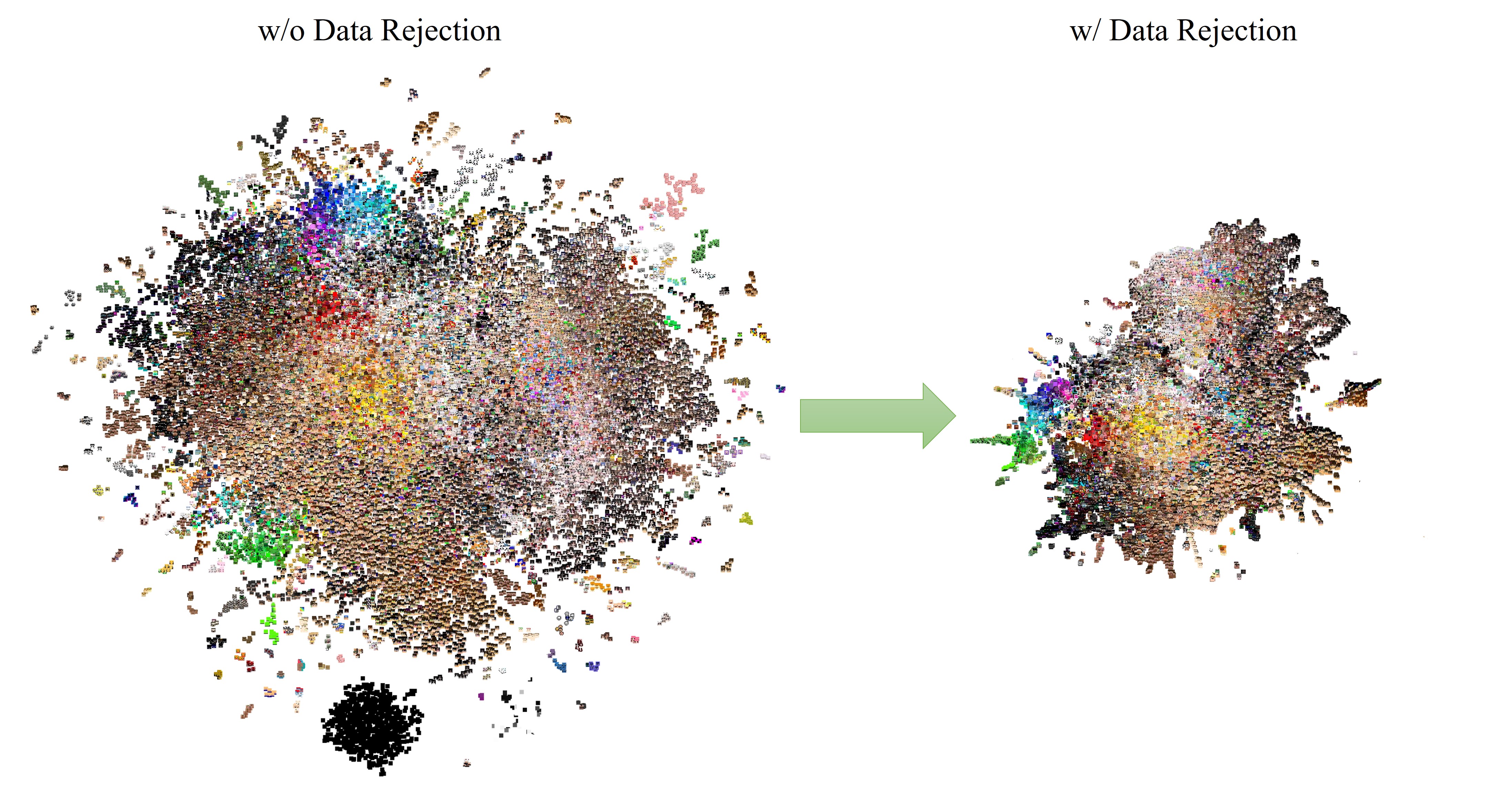}
    \hfil
\caption{Dataset refinement result.}	
\label{fig:supple_01}    
\end{figure}

Based on open-dataset, we further collect texture dataset to cover the unique or hand-crafted texture as many as possible. To elaborate the training dataset, data refinement process is conducted: (a) reject low standard deviation, (b) reject meaningless pattern image like chessboard, (c) reject monochromatic image. Total refined dataset include about 35K textures. Result about dataset refinement is shown in Fig. \ref{fig:supple_01}.

\section{Additional experiments} 

\begin{figure}[!htbp]
\centering
    \includegraphics[width=0.8\linewidth]{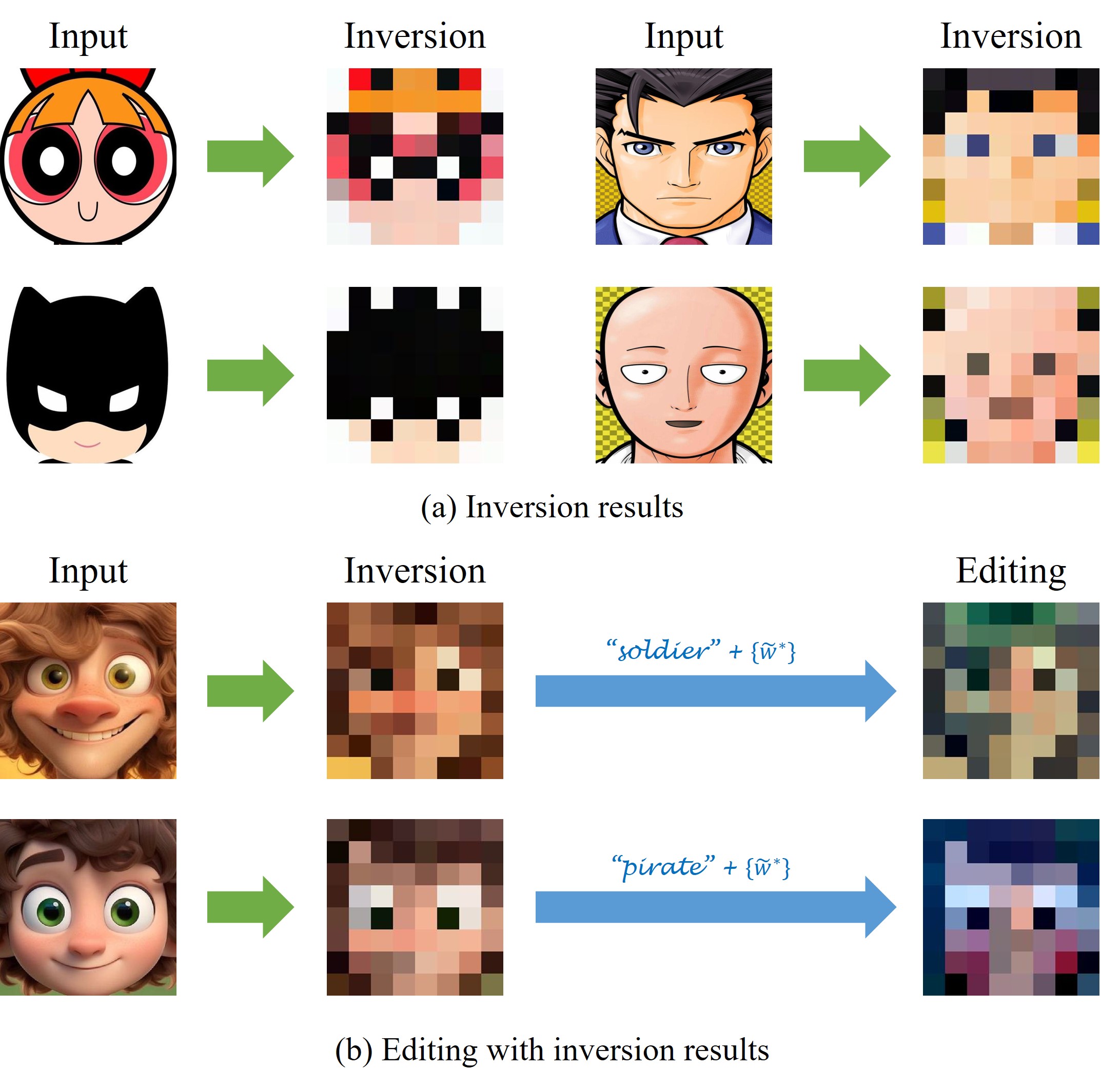}
    \hfil
\caption{Additional results with famous animation characters.}	
\label{fig:supple_02}    
\end{figure}

\begin{figure}[!htbp]
\centering
    \includegraphics[width=\linewidth]{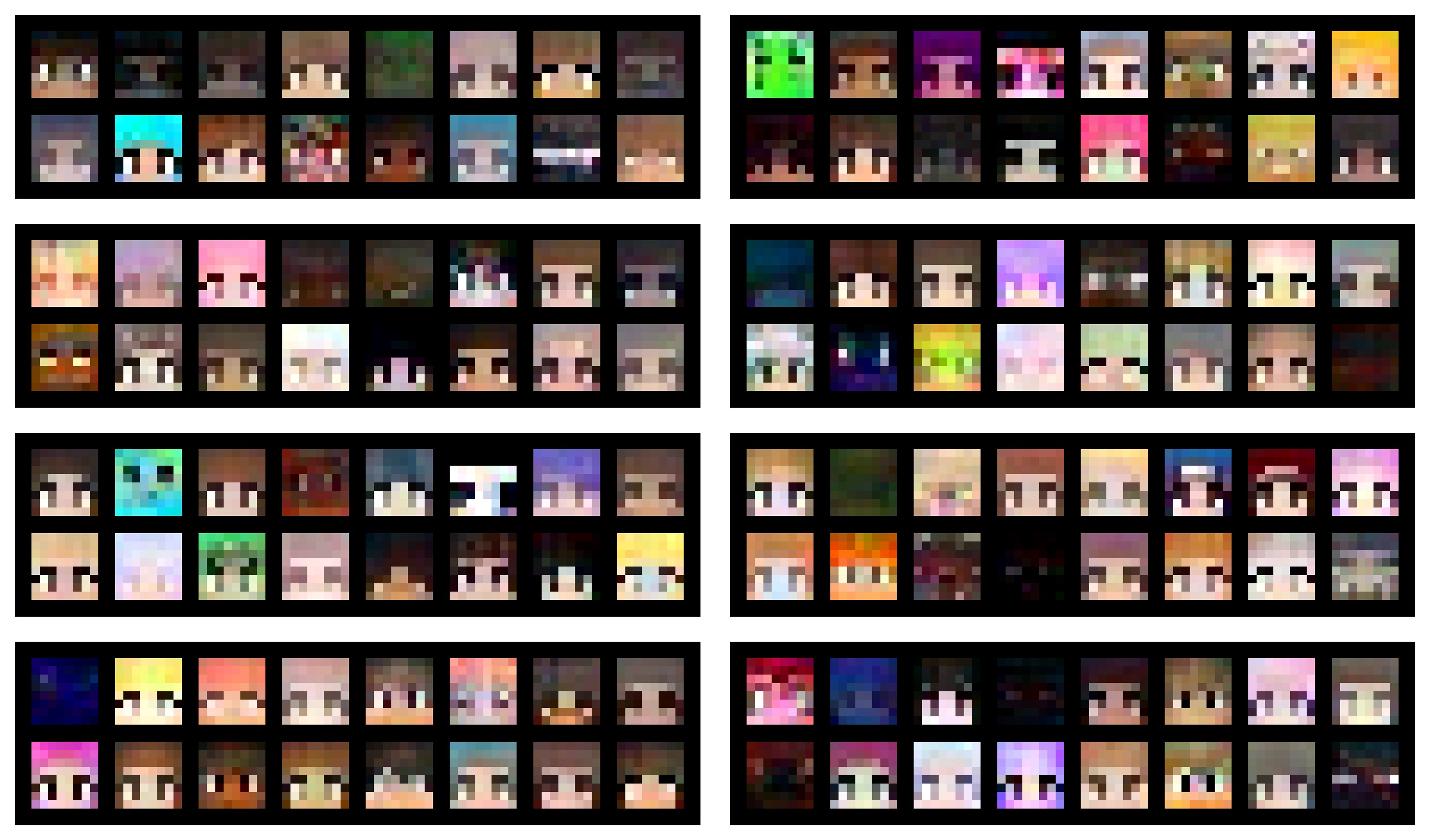}
    \hfil
\caption{Random generated texture from learned distribution.}	
\label{fig:supple_03}    
\end{figure}

To depict a character in popular or celebrated animation, we inverse and edit non-photorealistic image. As shown in Fig. \ref{fig:supple_02}, fine-level character faces (Fig. \ref{fig:supple_02} (b)) are easily collapsed while losing their detailed information. Simple or coarse-level character faces without high-frequency details are relatively preserved compared to aforementioned one, it often shows an unsatisfactory appearance by inversion process. It relies heavily on user-provided image sample that can come from a wide variety of domains including different rendering styles, structures, color distributions, and so on. In addition, we perform random generation from our learned distribution as shown in Fig. \ref{fig:supple_03}.

\section{In-Game screenshot} \label{appendix}

\begin{figure}[!htbp]
\centering
    \includegraphics[width=\linewidth]{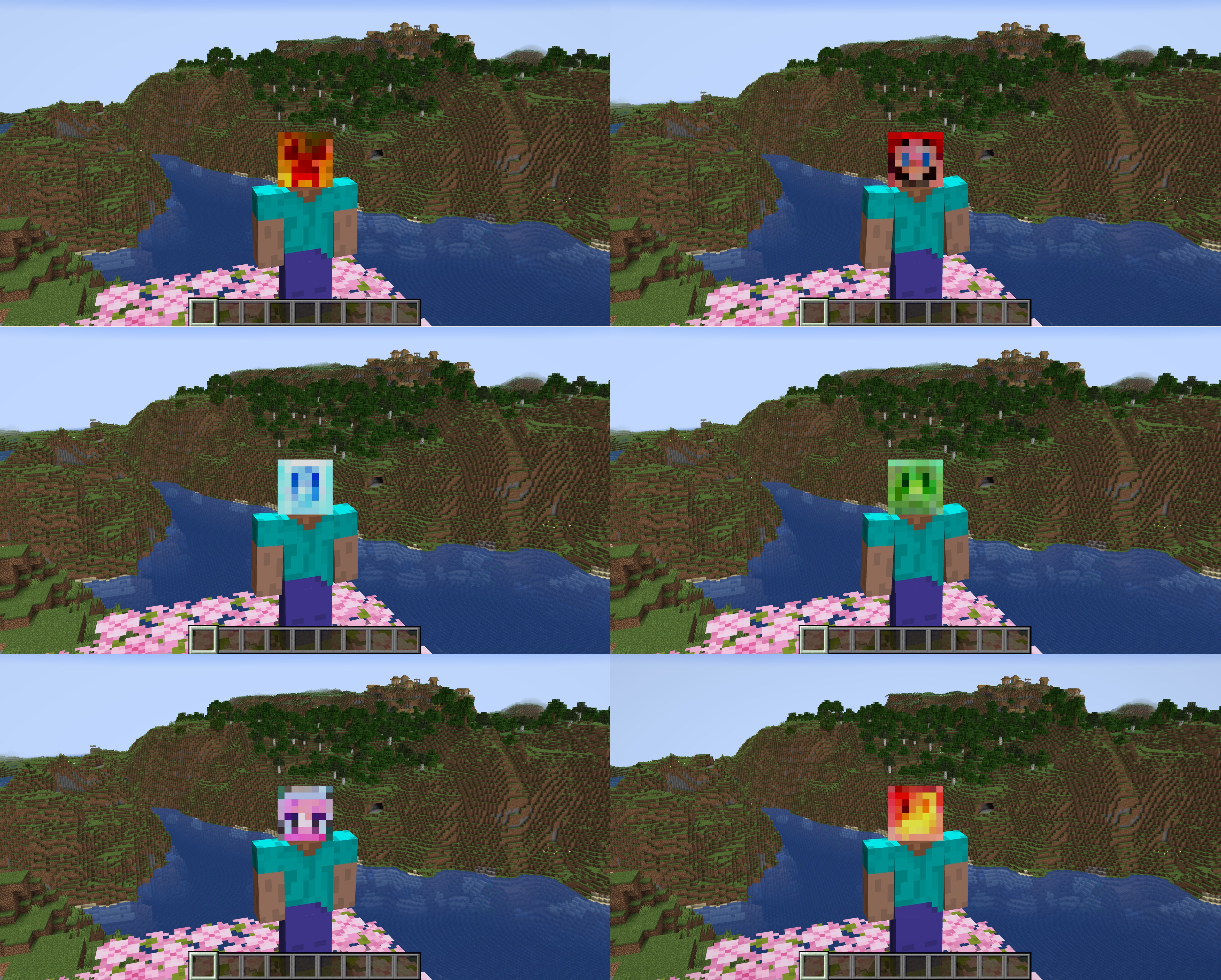}
    \hfil
\caption{In-Game screenshots using our edited face texture.}	
\label{fig:supple_04}    
\end{figure}

In this section, we showcase the overall screenshot images using our results as shown in Fig. \ref{fig:supple_04}

\section{Future work}

This work aims to generate character texture for in-game application. In Minecraft world, virtual character include face, body texture. For entire texture generation, we know that our system need to generate all the texture not only face but also body and others. To this end, we are continuing our \textit{Minecraft-ify} research project to cover this issue in additional method. Target goal of future work may include generating face, body, and accessories.

\end{document}